\documentclass[sigconf]{acmart}
\makeatletter
\def\@ACM@checkaffil{% Only warnings
    \if@ACM@instpresent\else
    \ClassWarningNoLine{\@classname}{No institution present for an affiliation}%
    \fi
    \if@ACM@citypresent\else
    \ClassWarningNoLine{\@classname}{No city present for an affiliation}%
    \fi
    \if@ACM@countrypresent\else
        \ClassWarningNoLine{\@classname}{No country present for an affiliation}%
    \fi
}
\makeatother
\AtBeginDocument{%
  \providecommand\BibTeX{{%
    \normalfont B\kern-0.5em{\scshape i\kern-0.25em b}\kern-0.8em\TeX}}}

\copyrightyear{2023} 
\acmYear{2023} 
\setcopyright{rightsretained} 
\acmConference[SEC '23]{The Eighth ACM/IEEE Symposium on Edge Computing}{December 6--9, 2023}{Wilmington, DE, USA}
\acmBooktitle{The Eighth ACM/IEEE Symposium on Edge Computing (SEC '23), December 6--9, 2023, Wilmington, DE, USA}
\acmDOI{10.1145/3583740.3628422}
\acmISBN{979-8-4007-0123-8/23/12}

\begin{document}

\title{Poster: Real-Time Object Substitution for Mobile Diminished Reality with Edge Computing}

\author{Hongyu Ke and Haoxin Wang}
\affiliation{%
  \institution{Georgia State University}
  % \city{Atlanta}
  % \state{Georgia}
  % \country{USA}
  % \postcode{30302}
}
\email{{hke3, haoxinwang}@gsu.edu}

% \author{Hongyu Ke}

% \affiliation{%
%   \institution{Georgia State University}
%   \city{Atlanta}
%   \state{Georgia}
%   \country{USA}
%   \postcode{30302}
% }
% \email{hke3@gsu.edu}

% \author{Haoxin Wang}
% \affiliation{%
%   \institution{Georgia State University}
%   \city{Atlanta}
%   \state{Georgia}
%   \country{USA}
%   \postcode{30302}
% }
% \email{haoxinwang@gsu.edu}

\renewcommand{\shortauthors}{Ke and Wang}

%%
%% The abstract is a short summary of the work to be presented in the
%% article.
\begin{abstract}
Diminished Reality (DR) is considered as the conceptual counterpart to Augmented Reality (AR), and has recently gained increasing attention from both industry and academia. Unlike AR which adds virtual objects to the real world, DR allows users to remove physical content from the real world. When combined with object replacement technology, it presents an further exciting avenue for exploration within the metaverse. Although a few researches have been conducted on the intersection of object substitution and DR, there is no real-time object substitution for mobile diminished reality architecture with high quality. In this paper, we propose an end-to-end architecture to facilitate immersive and real-time scene construction for mobile devices with edge computing.
\end{abstract}




\begin{teaserfigure}
  \includegraphics[width=0.98\textwidth]{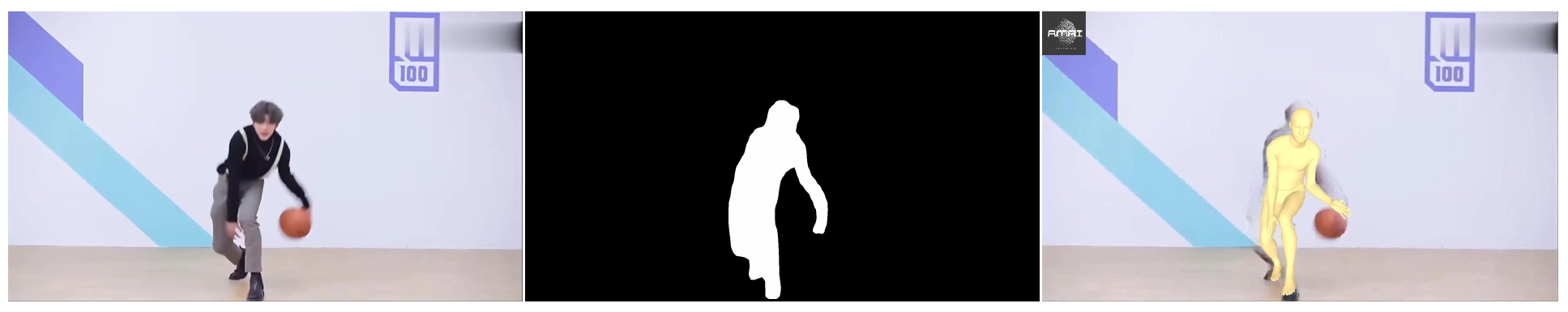}
  \vspace{-0.15in}
  \caption{Object substitution for diminished reality}
  \label{fig:cxk}
\end{teaserfigure}

\maketitle

\section{Background and Motivation}

Powered by augmented and virtual reality (AR/VR), the metaverse has materialized from the realms of science fiction in stages, and it can be expected to be well developed by 2040 \cite{anderson2022metaverse}. In AR, virtual objects are overlaid onto real world to augment users' perception of the world. These virtual objects provide relevant personalized information according to use cases and remaining consistent with the real world, creating an illusion of seamless blending. For example, the Meta-empowered advanced driver assistance system (ADAS) \cite{wang2023metamobility} provides additional information of neighboring drivers through multiple sources of data by displaying on the ego vehicle’s windshield as the augmented reality based head-up display to further enhance driving safety. This illustrates that the seamless integration of the virtual and real worlds, coupled with the consistent accessibility of digital data, are promising.

However, current AR primarily focuses on placing additional virtual objects to the real world to craft a seamless experience. Diminished Reality (DR), a class of augmentations, provides a contrasting approach by removing real world contents from users’ perception \cite{cheng2022towards}. This ability to selectively shape experiences makes the metaverse a more adaptable and customized digital frontier. Drawing from this and combining with object substitution, as showed in Fig. \ref{fig:cxk}, the metaverse offers a plethora of scenarios, such as a science-fiction scene where a pedestrian playing basketball is substituted with a cyberpunk-styled element. Beyond that, as the pedestrian crossing the street, cyberpunk-styled vehicles pass by him. 

While it is possible to directly overlay virtual objects onto their real world counterparts, they can be constrained by the shape and size of the original objects if the desired virtual objects are smaller than the original ones, and there can be mismatches or ambiguous occlusion relationships. For example, a pedestrian standing behind a light pole is substituted by an avatar. Moreover, the task of object substitution in DR demands a sophisticated level of scene understanding that draws on significant advances in machine perception. Creating such novel and real-time experiences from scratch is a considerable challenge.

In this paper, we propose an end-to-end system for real-time object substitution in mobile DR with edge computing. This system ensures the capability to create science-fiction futuristic scenes using mobile devices in real-time and enables users to have immersive interactions with virtual objects embedded in the real world. Our system complements traditional augmented reality where additional virtual objects nearly coexist with real ones. With our system, users can map poses of real objects to corresponding virtual ones, allowing the virtual objects to inherit semantics from the real world and enhancing interaction to a great extent.

\begin{figure}[t]
%left bottom right top
\centering
\includegraphics[width=0.48\textwidth]{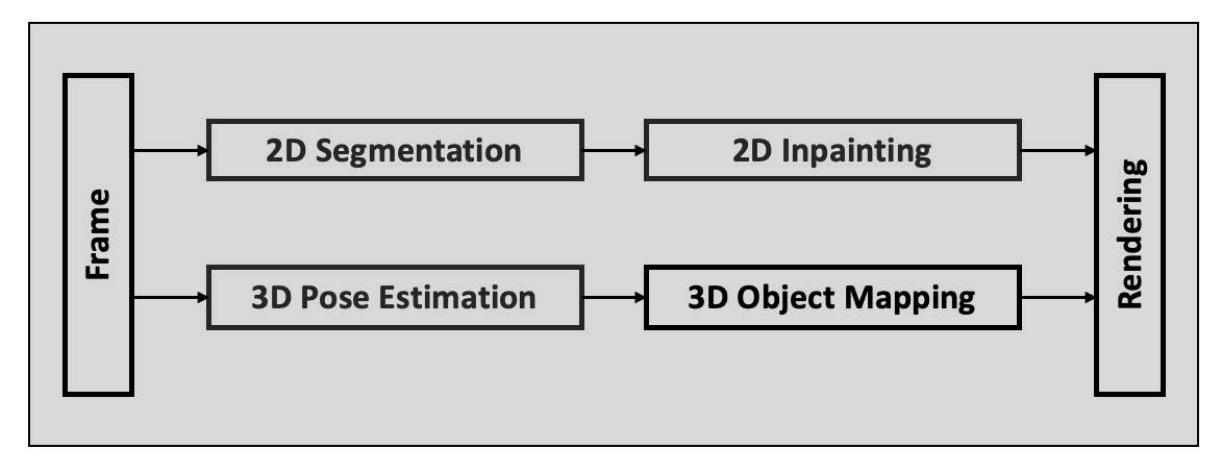}
\centering
\vspace{-0.2in}
\caption{Two major parallel pipelines of the proposed end-to-end system for real-time object substitution.}
\label{fig:structure}
\centering
\end{figure}

\begin{figure}[t]
%left bottom right top
\centering
\includegraphics[width=0.48\textwidth]{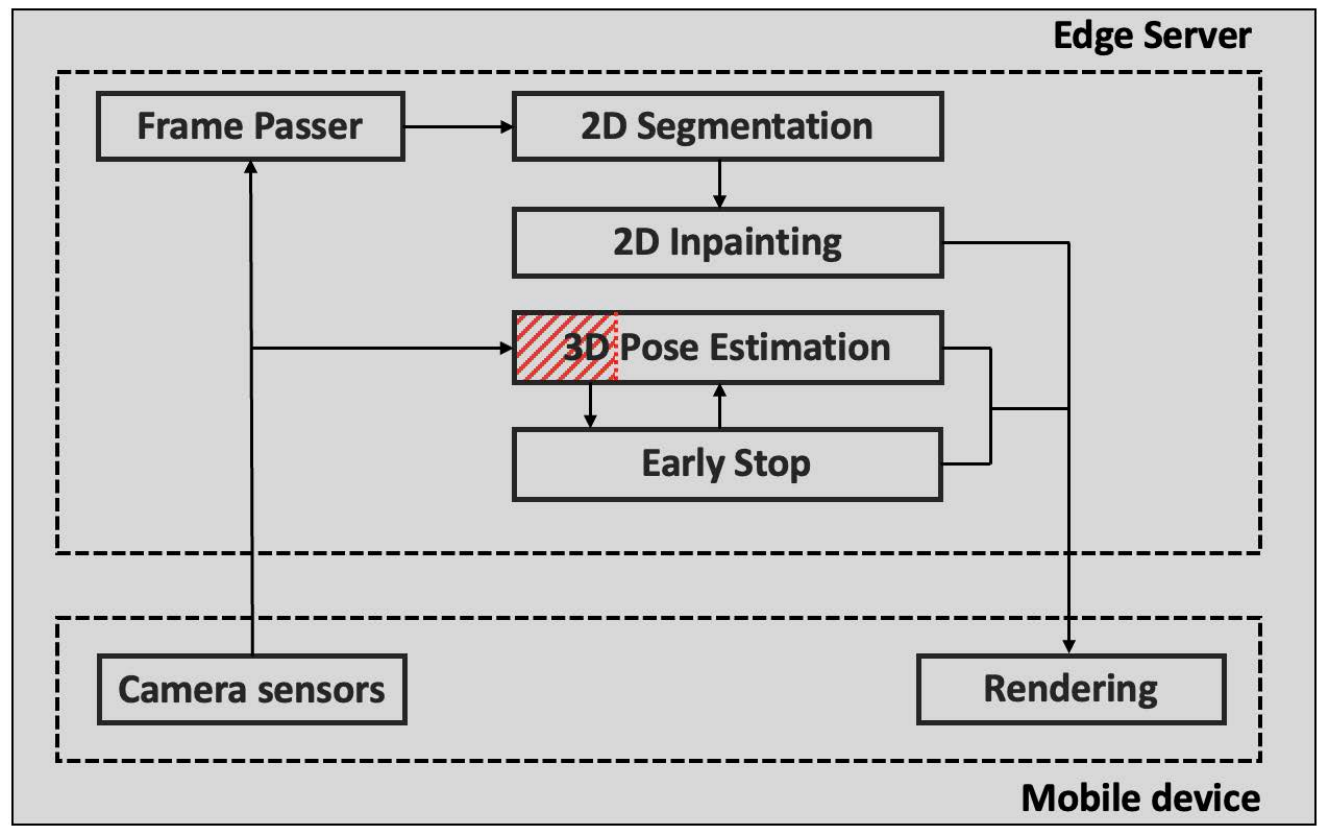}
\centering
\vspace{-0.2in}
\caption{Overview of the end-to-end system for real-time object substitution with edge computing.}
\label{fig:archecture}
\centering
\end{figure}

\section{Initial Design}

Inspired by \cite{kari2021transformr}, our proposed system consists of two major parallel pipelines, as shown in Fig. \ref{fig:structure}. A 2D pipeline is designed to achieve DR and a 3D pipeline is responsible for performing object substitution.  

The 2D pipeline comprises two major components: 2D segmentation and 2D inpainting. To achieve DR in a scene, a hole is created at the location of removed objects using inpainting, and identified through instance segmentation. This approach ensures the system works effectively, without previously captured images of the background, allowing for the deployment without constraints from environmental settings. 

The 3D pipeline comprises two major components: 3D pose estimation and 3D object mapping. To achieve object substitution, a primary goal is to render virtual objects into the scene in positions and orientations that mirror those of their physical counterparts. This requires to estimate poses of physical objects and mapping poses onto the virtual objects subsequently. For 3D pose estimation, challenges such as occlusions and clutter backgrounds in the environment \cite{sarafianos20163d} can hinder accurate pose estimation. One of the most effective current methods to address these challenges is to use multi-camera systems or combining data from multiple sensors. These approaches are compatible with many recent mobile devices, such as Microsoft HoloLens2 and most recent smartphones.

After processing the data through both the 2D and 3D pipelines, the inpainted frame and 3D pose estimation are obtained. By running the 3D scene rendering, these components work in tandem to provide a seamless integration of virtual objects into the real scene.

\section{Challenges and Future Work}
\label{sc:Results}

There are several challenges associated with real-time object substitution for mobile DR with edge computing system.
\begin{enumerate}
    \item Even when we offload computation-intensive tasks to the edge server, such as segmentation, inpainting, and pose estimation, it remains difficult to reduce the latency to a senseless level such as 30fps.
    \item There is a lack of tools and metrics to measure scene quality, making it challenging to quantitatively evaluate object substitution for DR.
    \item Current system is primarily designed for the single user scenarios, it fails to consider the field of collaborative AR/DR where multiple users interact within the same environment. 
\end{enumerate}

Our future work will be mainly based on those challenges. At the moment, we plan to implement this system in an edge-server-based architecture and offload the computation-intensive operations to the server, as showed in Fig. \ref{fig:archecture}. Based on two major parallel pipelines, we suppose to immediately send the image and construct the virtual objects on the mobile device. Inspired by \cite{wu2020emo}, we also plan to employ two system technique modules, the frame passer and the early stop. These will exploit the seen background and the feature similarity of continuous frames, respectively, for the sake of accelerating processing and saving resources.
The frame passer, applied before the 2D segmentation, selectively forwards frames to the 2D segmentation module based on the recording caches from the cameras. If consecutive frames capture the background information of the location where the target object is positioned, this component processes the current frame using a simplified method and bypass the rest of 2D pipeline. Its aim is to minimize computational overhead while ensuring unseen scenes are not missed.
The early stop, which obtains intermediate results from 3D pose estimation, is to assess whether two consecutive inputs are similar in terms of their distinctive pose features. If they are, this component allows the current input to bypass the remaining stages of the pose estimation process and assigns it the same pose as the previous input frame.

\section{Conclusion}
In this paper, we proposed an end-to-end system for real-time object substitution with edge computing. Research motivation and initial system design were presented to support the rationale behind our approach and highlight the potential benefits this system brings to the field of metaverse.

% \bibliographystyle{ACM-Reference-Format}
% \bibliography{sample-base}
\bibliographystyle{unsrt}
\bibliography{references}

\end{document}